\documentclass{article}

\usepackage{microtype}
\usepackage{graphicx}
\usepackage{subfigure}

\usepackage{booktabs} 

\usepackage{hyperref}

\usepackage[accepted]{icml2018}

\icmltitlerunning{Planning in Dynamic Environments with Conditional Autoregressive Models}

\begin{document}

\twocolumn[
\icmltitle{Planning in Dynamic Environments with Conditional Autoregressive Models}

\icmlsetsymbol{equal}{*}

\begin{icmlauthorlist}
\icmlauthor{Johanna Hansen*}{mcgill}
\icmlauthor{Kyle Kastner*}{udem}
\icmlauthor{Aaron Courville}{udem,cifar}
\icmlauthor{Gregory Dudek}{mcgill}
\end{icmlauthorlist}

\icmlaffiliation{mcgill}{Mobile Robotics Lab, School of Computer Science, McGill University, Montr\'eal, Qu\'ebec, Canada}
\icmlaffiliation{udem}{Montr\'eal Institute for Learning Algorithms (MILA), Universit\'e de Montr\'eal, Montr\'eal, Qu\'ebec, Canada}
\icmlaffiliation{cifar}{CIFAR Fellow}

\icmlcorrespondingauthor{Johanna Hansen}{johanna.hansen@mail.mcgill.ca}

\icmlkeywords{Machine Learning, Planning, ICML}

\vskip 0.3in
]



\printAffiliationsAndNotice{\icmlEqualContribution} 

\begin{abstract}

We demonstrate the use of conditional autoregressive generative models \cite{pixelcnn} over a discrete latent space \cite{vqvae} for forward planning with MCTS. In order to test this method we introduce a new environment featuring varying difficulty levels, along with moving goals and obstacles. The combination of high-quality frame generation and classical planning approaches nearly matches true environment performance for our task, demonstrating the usefulness of this method for model-based planning in dynamic environments. 

\end{abstract}
\section{Introduction}
Planning agents find actions at each decision point by considering future scenarios from their current state against a model of their world \cite{rrt, uct_csaba, dstar, anytime_dynamic}. 
Though typically slower at decision-time than model-free agents, agents which use planning can be configured and tuned with explicit constraints. Planning based methods can also reduce the compounding of errors for sequential decisions by directly testing long term consequences from action choices, balancing exploitation and exploration, and generally limiting 
issues with long-term credit assignment. 

Model-free reinforcement learning approaches are often sample inefficient, requiring millions of steps to jointly learn environment features and a control policy. Agents which employ decision-time planning techniques, on the other hand, do not explicitly require any training prior to decision time. 
However, to perform well, planning-based agents need a very accurate future model of their environment for evaluating actions. A perfect model of the future to perform forward planning is usually not possible outside of computer games or simulations. In this paper, we demonstrate how we can leverage recent improvements in generative modeling to create powerful dynamics models that can be used   for forward planning. 

In this paper we discuss an approach for learning conditional models of an environment in an unsupervised manner, and demonstrate the utility of this model for use with decision-time planning in a dynamic environment. Autoregressive models have shown great results in generating raw images, video, and audio \cite{pixelcnn, wavenet,  videopixelnetworks}, but have generally been considered too slow for use in decision making agents \cite{learning_and_querying}. However, in \cite{vqvae}, the authors show that these autoregressive models can be used as a generative prior over the latent space of discrete encoder/decoder models. Operating over these concise latent representations of the data instead of pixel-space greatly reduces the time needed for generation, making these models feasible for use in decision-making agents. 



\section{Background}
Learning accurate models of the environment has long been a goal in model-based reinforcement learning and unsupervised learning. Recent work has shown the power of learning action-conditional models for training  decision-making agents with  perceptual models \cite{world_models, on_learning_to_think,learning_and_querying,oh2015, graves13} and combining planning and with environment models \cite{predictron, amys_composable_planning,razvan, mcts_nets,thinking_fast_and_slow,mcts_nets}.  

 For real-world agents, semantic information is often more relevant than perceptual input for task performance and planning \cite{NextSegmPredICCV17}.
Our experimentation over semantic space shows that for our task, a VQ-VAE model greatly outperforms VAE \cite{VAE} reconstructions. Instead of assuming normally distributed priors and posteriors as in a typical VAE architecture, VQ-VAEs learns categorical distributions in the latent space where the samples from the distributions are indexes to an embedding table. Van den Oord et al. \cite{vqvae} demonstrates the benefits of learning action-condition and action-independent forward predictions over VQ-VAE latent space. We build upon this work by combining it with a classical method for planning in order to navigate in an environment with numerous dynamic obstacles and a moving target.



We test our forward-model with a powerful anytime planning method, Monte-Carlo Tree Search (MCTS) \citep{uct_csaba}. Given an accurate representation of the future and sufficient time to compute, MCTS performs well \citep{mspacmanmcts}, even when faced with large state or action spaces. MCTS works by \emph{rolling out} many sequences of actions possible future scenarios to acquire an approximate (Monte Carlo) estimate of the value of taking a specific action from a particular state. For a full overview of MCTS and its many variants, please refer to \cite{MCTSSurvey}.  MCTS has been used in a wide variety of search and planning problems where a model of the world is available for querying  \cite{AlphaGOSilver_2016, MCTS_RAM, ATARI, classical_planning_atari, mcts_atari}. The performance of MCTS is critically dependent on having an accurate forward model of the environment, making it an ideal fit for testing our autoregressive conditional generative forward model. 


\section{Experiments}

We consider a fully-observable task in which an agent must navigate to a dynamic goal location without contact with moving obstacles. 
At each time step $t$, the agent realizes an observation $o_t$ and must execute an action $a_t$.  In our experiments, the observation is an image constituting the full view of an action-independent, two-dimensional environment. The action space consists of $8$ actions, where each action moves a fixed amount in a specific direction, diagonal included. We learn a conditional forward model of this environment as described in Section \ref{sec:genmodel} and query at decision time for action selection with MCTS.

Our problem is similar to those faced by autonomous underwater vehicles (AUVs) navigating in a busy harbor while try to avoid traveling underneath passing ships \citep{risk_aware_shipping_channel}. In order to successfully accomplish this tasks, the robot needs reliable dynamics models of the obstacles (ships) and goals in the environment so it can plan effectively against a realistic estimate of future states. 

\begin{figure}[h!]
         
\centering
   \begin{minipage}[b]{\linewidth}
        \includegraphics[trim={3cm 0 3cm 0},clip,width=\textwidth]{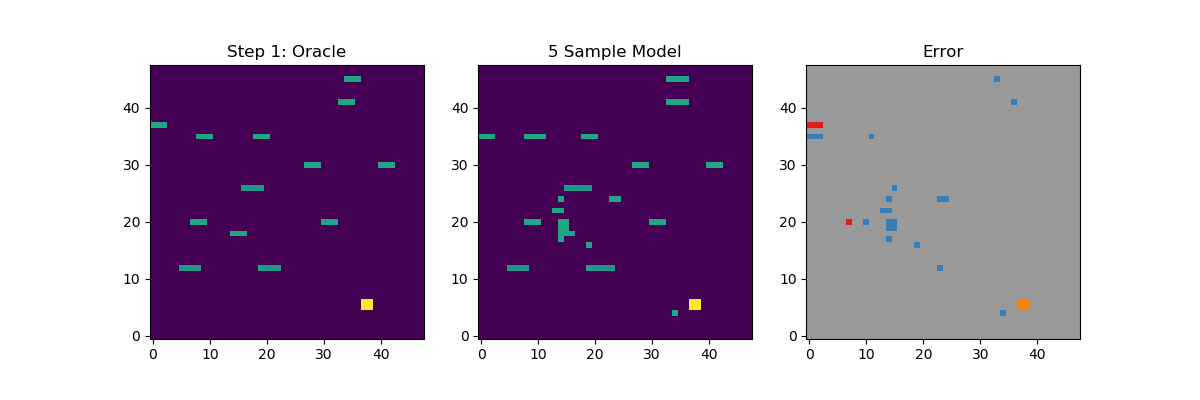}
     
    \end{minipage} %
         
    \vspace{-5.00mm}
    
	\centering
    \begin{minipage}[b]{\linewidth}   
        \includegraphics[trim={3cm 0 3cm 0},clip,width=\textwidth]{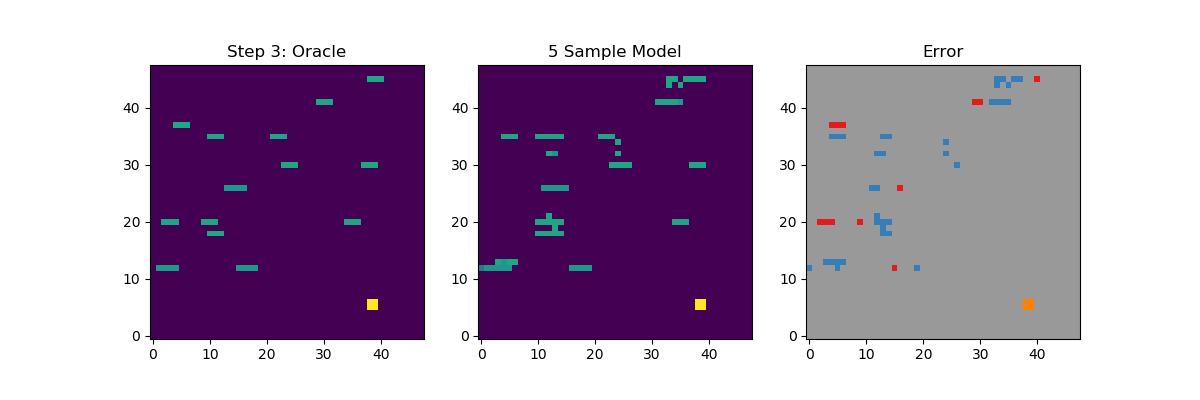}
      
    \end{minipage}
           
    \vspace{-5.00mm}
    
    \centering
    \begin{minipage}[b]{\linewidth}
       \includegraphics[trim={3cm 0 3cm 0},clip,width=\textwidth]{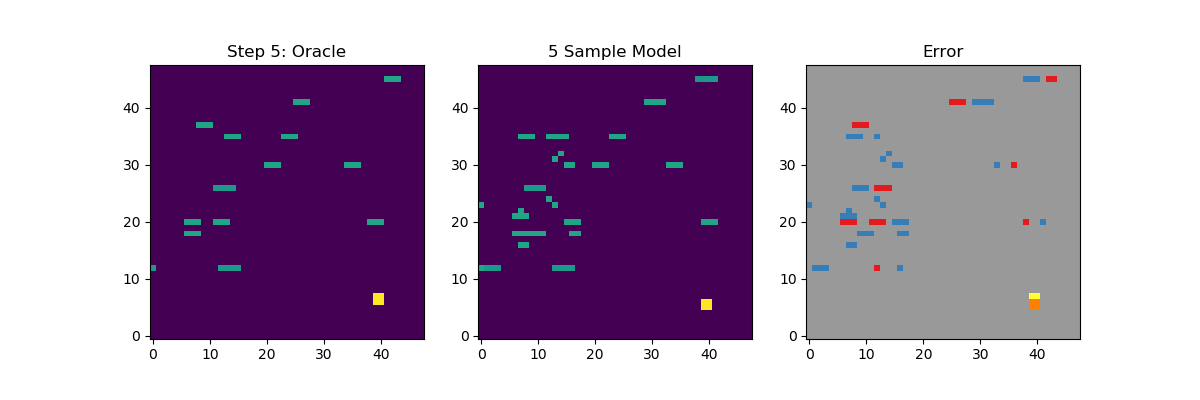}
     
    \end{minipage}
       
    \vspace{-5.00mm}
    
     \centering
    \begin{minipage}[b]{\linewidth}
        \includegraphics[trim={3cm 0 3cm 0},clip,width=\textwidth]{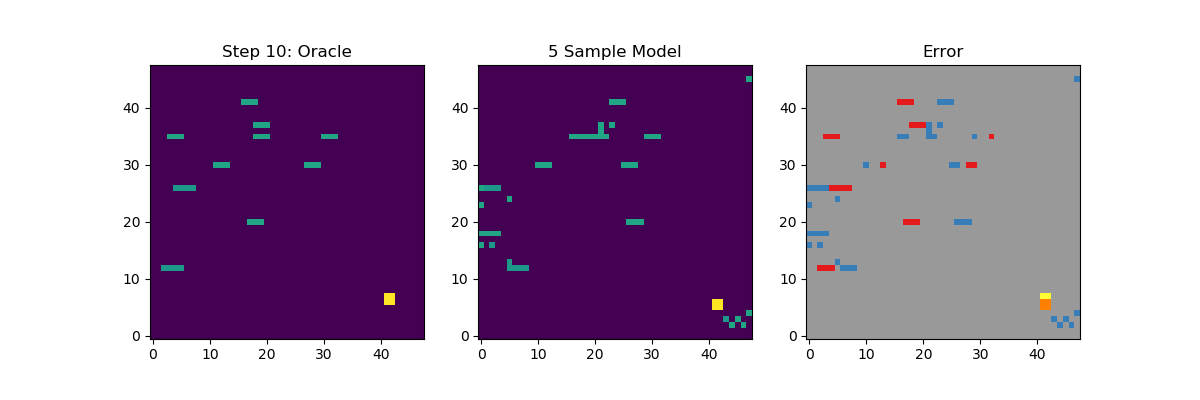}
         
    \end{minipage}
    
    \vspace{-5.00mm}
    
\caption{This figure illustrates forward rollout steps by the oracle (left column), our $5$ sample model (middle column), and the error in the model (right column). The number of steps from the given  state $t$ is indicated in the oracle plot's title.   In the first two columns, free space is violet, moving obstacles are cyan, and the goal is  yellow. In the third column, we illustrate obstacle error in the model as follows: false negatives (predicted free space where there should be an obstacle) are red and false positives (predicted obstacle where there was free space) are blue. The true goal is plotted in yellow and the predicted goal is plotted in orange (perfect goal prediction is orange). }
\label{fig:compare}
\vspace{-4.00mm}
\end{figure}

\subsection{Environment Description}
We introduce a navigation environment (depicted in the first column of Figure  \ref{fig:compare}) which consists of a configurable world with dynamic obstacles and a moving goal. Movement about the environment is continuous, but collision and goal checking is quantized to the nearest pixel. In each episode the $1\times1$ size agent and $2\times2$ size goal are initialized to a random location, and the goal is given a random vector direction and a fixed velocity. The agent must then reach the moving goal within a limited number of steps without colliding with an obstacle.  At each timestep the agent has the choice of $8$ actions. These actions indicate one of $8$ equally spaced angles and a constant speed. In these experiments, we test two agents, one at $0.5$ pixels per timestep ($1\times$ goal agent) and one agent at $1$ pixels per timestep ($2\times$ goal agent). The goal moves about the  environment at a fixed random angle and fixed speed of $0.5$ pixels per timestep. The goal also reflects off of world boundaries, making good modeling of goal dynamics important to success.

The environment is divided into obstacle \emph{lanes} which span the environment horizontally. At the beginning of each episode, the lanes are randomly assigned to carry $1$ of $5$ classes of obstacles and a direction of movement (left to right or right to left). Each obstacle class is parameterized by a color and a distribution which describes average obstacle speed and length. Obstacles maintain a constant speed after entering the environment, pass through the edges of the environment, and are deleted after their entire body exits the observable space. The number of obstacles introduced into the environment at each timestep is controlled by a Poisson distribution, configured by the \emph{level} parameter. For the results reported in this paper we set the level to $6$, however there is support for a variety of difficulty settings. At each time step, the observation consists of the agent's current location and the full quantized pixel space including the goal and obstacles.  

An agent receives a reward of $+20$ for entering the same pixel-space as the goal and a $-20$ reward for entering the same pixel-space as an obstacle. Both events cause the episode to end. The agent has a limited number of actions before the game times out, resulting in a reward of $0$. This step limit is dependent on the speed of the agent and the size of the grid. For these experiments, the $2\times$ agent has $203$ steps and the $1\times$ agent has $407$ steps before the game ends.

A key component which makes our approach computationally feasible is that the environments of concern are \emph{not} action conditional, meaning dynamics in the world continue regardless of what actions are chosen. This means that generated future frames can be shared across all rollouts in MCTS, greatly reducing the overall sample cost for the autoregressive model. Combined with the speed improvements from generating in a compressed space given by VQ-VAE, forward generation can be accomplished in reasonable time. It is also possible to take a similar approach in action-conditional spaces, but this would increase the number of needed generations from the model during MCTS rollout by a large amount.

\subsection{Model Description}
\label{sec:genmodel}
We utilize a two-phase training procedure on the agent-independent, $48\times48\times1$ environment described in the previous section. First we learn a compact, discrete representation (denoted $Z$) of individual pixel-space frames with a VQ-VAE  model \cite{vqvae} with discretized logistic mixture  likelihood \cite{pixelcnnpp} for the reconstruction loss. In the second stage, an autoregressive generative model, a conditional gated PixelCNN \cite{pixelcnn} is trained to predict one-step ahead $Z$ representations of sequential frames when conditioned on previous $Z$ representations. To introduce Markovian conditions, the conditional gated PixelCNN is fed a spatial conditioning map of $4$ past $Z$ encodings, in addition to the current step. The resulting PixelCNN learns a model corresponding to $p(Z_{t_{i, j}} | Z_{t_{<i, <j}}, Z_{t-1}, Z_{t-2}, Z_{t-3}, Z_{t-4})$, where each dimension $(i, j)$ of $Z_t$ is conditioned on all valid dimensions relative to the current position via autoregressive masking, and also conditioned on the previous $4$ frames by a spatial conditioning map \citep{pixelcnn} which is fed as input. Combined with the previously trained VQ-VAE decoder this results in a model which generates $1$ frame ahead, given $4$ previous frames. It is possible to generate an arbitrary number of frames forward given an initial $4$ frames, by chaining $1$ step generations though we expect results to degrade as forward trajectory lengths increase.  

\begin{table*}[h!]
\caption{Performance Comparison over 100 Episodes}
\label{tab:tab}
\setlength{\tabcolsep}{2pt}
\vskip 0.15in
\begin{center}
\begin{small}
\begin{sc}
\begin{tabular}{||l||cccc||cccc||cccc||cccc||}
\toprule
Rollout Steps & 
\multicolumn{4}{|c||}{1} &
\multicolumn{4}{|c||}{3} &
\multicolumn{4}{|c||}{5} &
\multicolumn{4}{|c||}{10} \\
\toprule
Technique  &G&T&D&S &G&T&D&S &G&T&D&S &G&T&D&S\\ 
\midrule

2$\times$Oracle  &100&0&0&34x$\pm$17 &100&0&0&36$\pm$18 &100&0&0&45$\pm$28 &100&4&0&68$\pm$49\\
2$\times$Mid &78&0&22&33$\pm$17  &88&0&12&40$\pm$18 &91&2&7&65$\pm$40 &52&25&23&111$\pm$67 \\
2$\times$5 Samples  &84&0&16&34$\pm$17 &
\boldmath{$94$}&1&5&46$\pm$27 &89&5&6&75$\pm$51 &55&23&22&112$\pm$70 \\
2$\times$10 Samples  &85&0&15&35$\pm$18  &88&0&12&46$\pm$26 &89&9&2&76$\pm$56 &55&31&14&124$\pm$68 \\
\midrule
1$\times$Oracle   &72&25&3&187$\pm$151  &67&32&1&209$\pm$154  &60&40&0&224$\pm$64  &66&34&0&216$\pm$156 \\
1$\times$5 Samples   &31&3&55&$97\pm107$ 
&46&21&33&196$\pm$153  &41&3&27&259$\pm$155 
&39&46&15&294$\pm$143 \\
\bottomrule
\end{tabular}
\end{sc}
\end{small}
\end{center}
\vskip -0.1in
\caption{This table compares agents using MCTS for forward planning on varying models (oracle and ours with varying levels of sampling from the generative model), rollout lengths (1, 3, 5 and 10), and agent speed (2X agents are twice as fast as the goal and 1X agents  are the same speed as the goal). All agents were tested over the same set of $100$ random episodes, with MCTS performing $100$ rollouts at each decision time. The values in columns \emph{G}, \emph{T}, and \emph{D} stand for the number of games in which the described agent reached the goal (\emph{G}), ran out of time before reaching the goal (\emph{T}), or died (\emph{D}) by running into an obstacle. The \emph{S} column describes the number of steps completed on average by an agent, calculated only from episodes in which the agent avoided dying (smaller is better), along with the standard deviation. When tested on the same episodes, a random agent reached the goal once at $2$X speed and never at $1$X speed. }
\end{table*}
\subsection{MCTS Planning}

Our MCTS agent is characterized by rollout length, number of rollouts, and temperature. We vary rollout length from $1$ to $10$, but hold the number of rollouts to $100$ and temperature to $0.01$ for all experiments. 
We also use a goal-oriented prior for node selection as described by prior work using PUCT MCTS \citep{Rosin2011,AlphaGoZero}. This prior biases tree expansion during rollouts such that actions in the direction of the predicted goal are more likely to be chosen. Adding goal information to the state has been found to improve agents in other scenarios \cite{sukhbaatar2017intrinsic}, and we found that this simple prior greatly improved performance compared to a uniform prior, resulting in shorter average rollout lengths.  


\subsection{Training}
The VQ-VAE encoder consists of $4$ strided convolutional layers with a kernel size $(4, 4)$ and sizes of $42$, $32$, $16$, $16$. The first $3$ layers have strides of $2$ and the last layer has a stride of $1$. This configuration compresses an input size of $48\times48\times1$ down to a $Z$ space of  $6\times6\times1$. For learning the vector quantization codebook, we set K=$512$, resulting in a compression of $\frac{48\times48\times3}{6\times6\times9}\approx21.3$ in bits over each frame, considering there are $6$ pixel-values used in the input image (requiring $2^3$ bits to encode minimally). The VQ-VAE decoder inverts this process using transpose convolutions, and appropriate stride values which mimic the decoder settings but in reverse order. 

Training was performed for $64$ epochs with a minibatch size of $32$ over $837,270$ example frames which were generated from running the environment. We use an Adam optimizer \citep{kingma2014adam} with the learning rate set to $1e-3$, and the discretized mixture of logistics loss \citep{pixelcnnpp}. From the trained VQ-VAE model, we generate a new dataset consisting of ordered $Z$ values given by our model over $3000$ previously unseen episodes which are each $407$ frames long. The PixelCNN \cite{pixelcnn} is trained over these generated $Z$s for $10$ epochs with a batch size of $64$. We employ categorical cross-entropy loss and the Adam optimizer (learning rate is set to $0.0003$) for predicting the discrete "label" of each $Z$ dimension. We condition each prediction on a spatial map consisting of the previous $4$ frame's $Z$s \citep{pixelcnn}. 

\section{Performance}
\label{sec:perf}
Our experiments (see Table \ref{tab:tab}) demonstrate the feasibility of using conditional autoregressive models for forward planning. Example playout gifs can be found in the code repository at  \url{https:github.com/johannah/trajectories}. We compare agents using  our forward model to an agent which has access to an oracle of the environment. The oracle agent is used as an upper-bound on performance, as although this perfect representation of the future environment is not available in realistic tasks it is the theoretical best we can expect generative model to do. In all of the compared models, we first use a mid point "average" estimate from the discretized mixture of logistics distribution, but in those denoted by \emph{sampled}, we also sample an additional $5$ or $10$ times from the model and take the pixel-wise max of the predicted obstacle values. We find this results in a more conservative, but noisier estimate of the car locations. We take the median location of goal estimates over all of the samples to set the directional MCTS prior. 

Errors in the forward predictions (see Figure \ref{fig:compare}) can cause the agents to make catastrophic decisions, resulting in lower performance when compared to the oracle. False negatives, in particular (shown in red in Figure \ref{fig:compare}), result in the agent mistaking an obstacle for free space. Some of these mistakes are unavoidable as we step farther from the given state as we can only model obstacles that are in the scene at the current time step. This characteristic limits the efficacy of the lengths we can model forward in time and is a phenomena also discussed in Luc et al. \citep{NextSegmPredICCV17}.  

Perhaps unsurprisingly, our results show that the faster ($2\times$) agent had an easier time reaching the goal before running out of time.  
Agents which utilize longer rollouts were likely hampered by our decision to hold the number of rollouts constant over all of our experiments. Overall, longer rollouts were more likely to die off in their future states and thus often failed to come up with aggressive paths. 

Each future timestep prediction with our VQ-VAE + PixelCNN takes approximately $0.4$ seconds on a TitanX-Pascal GPU. An average action decision with our best performing agent ($2\times$ $5$ Samples with $3$ step rollouts) takes approximately $1.7$ seconds. Beyond using VQ-VAE to reduce the input space to PixelCNN, no other methods for improving the speed of autoregressive generation were employed. Recent publications in this area \citep{parallelwavenet, efficientwavenet, fastpixelcnn} show massive improvements in generation speed for autoregressive models and are directly applicable to this work.



\section{Conclusion}
We show that the two-stage pipeline
of VQ-VAE \citep{vqvae} combined with a PixelCNN prior conditioned on previous frames captures important semantic structure in a dynamic, goal oriented environment. The resulting samples are usable for model-based planning with MCTS over generated future states. Our agent avoids moving obstacles and reliably intercepts a non-stationary goal in the dynamic test environment introduced in this work, demonstrating the efficacy of this approach for planning in dynamic environments.




\bibliography{bib}
\bibliographystyle{icml2018}

\end{document}